\documentclass[runningheads]{llncs}
\usepackage[utf8]{inputenc}

\usepackage{graphicx}
\usepackage{float}
\usepackage{soul,color}

\begin{document}
\title{On the performance of deep learning models for time series classification in streaming}
\titlerunning{Deep learning models for time series classification in streaming}
\author{Pedro Lara-Benítez\inst{1}%\orcidID{0000-0003-0457-8099}
\and
Manuel Carranza-García\inst{1}%\orcidID{0000-0002-4729-8604}
\and \\
Francisco Martínez-Álvarez\inst{2}%\orcidID{0000-0002-6309-1785}
\and José C. Riquelme\inst{1}%\orcidID{0000-0002-8243-2186}
}
\authorrunning{P. Lara-Benítez et al.}
\institute{Division of Computer Science, University of Sevilla, ES-41012 Seville, Spain \and Data Science \& Big Data Lab, Pablo de Olavide University, ES-41013 Seville, Spain \email{plbenitez@us.es}\\}

\maketitle  % typeset the header of the contribution
\begin{abstract}
Processing data streams arriving at high speed requires the development of models that can provide fast and accurate predictions. Although deep neural networks are the state-of-the-art for many machine learning tasks, their performance in real-time data streaming scenarios is a research area that has not yet been fully addressed. Nevertheless, there have been recent efforts to adapt complex deep learning models for streaming tasks by reducing their processing rate. The design of the asynchronous dual-pipeline deep learning framework allows to predict over incoming instances and update the model simultaneously using two separate layers. The aim of this work is to assess the performance of different types of deep architectures for data streaming classification using this framework. We evaluate models such as multi-layer perceptrons, recurrent, convolutional and temporal convolutional neural networks over several time-series datasets that are simulated as streams. The obtained results indicate that convolutional architectures achieve a higher performance in terms of accuracy and efficiency.

\keywords{Classification \and Data Streaming \and Deep Learning \and Time Series.}
\end{abstract}
\section{Introduction}
Learning from data arriving at high speed is one of the main challenges in machine learning. Over the last decades, there have been several efforts to develop models that deal with the specific requirements of data streaming. Traditional batch-learning models are not suitable for this purpose given the high rate of arrival of instances. In data streaming, incoming data has to be rapidly classified and discarded after using it for updating the model. Predicting and training have to be done as fast as possible in order to maintain a processing rate close to real-time. Furthermore, the models have to be able to detect possible changes in the incoming data distribution, which is known as concept drift.

Despite the incremental learning nature of neural networks, there is little research involving deep learning (DL) models in the data streaming literature. Neural networks can adapt to changes in data by updating their weights with incoming instances. However, the high training time of deep networks presents challenges to adapt them to a streaming scenario. Very recently, our research group developed a deep learning framework for data streaming classification that uses an asynchronous dual-pipeline architecture (ADLStream) \cite{Lara:2020}. In this framework, training and classification can be done simultaneously in two different processes. This separation allows to use DL networks for data arriving at high speed while maintaining a high predictive performance. 

The aim of this study is to evaluate how different DL architectures perform on the data streaming classification task using the ADLStream framework. Despite the promising results presented in \cite{Lara:2020}, the experiments only considered convolutional neural networks, hence the suitability and efficiency of other types of deep networks is an area that has yet to be studied. In this work, we focus the experimental study on time-series data obtained from the UCR repository that have been simulated as streams. For this reason, we have designed DL models that are suitable for data having an inner temporal structure. The basic Multi-Layer Perceptron (MLP) is set as the baseline model and compared with other three architectures: Long-Short Term Memory network (LSTM), Convolutional Neural Network (CNN), and Temporal Convolutional Network (TCN). These models are evaluated in terms of accuracy and computational efficiency.

The rest of the paper is organised as follows: Section 2 presents a review on related work; Section 3 describes the materials used and the methodology; in Section 4 the experimental results obtained are reported; Section 5 presents the conclusions and future work.

\section{Related work}
Over the last decades, there have been several efforts to develop models that deal with the specific requirements of data streaming. Traditional batch-learning models are not suitable for this purpose given the high rate of arrival of instances. In data streaming, incoming data has to be rapidly classified and then discarded after using it for updating the learning model. Predicting and training have to be done as fast as possible in order to maintain a processing rate close to real-time. Furthermore, the models have to be able to detect possible changes in the incoming data distribution, which is known as concept drift\cite{Anderson:2019}. 

One of the most popular approaches has been to develop incremental or online algorithms based on decision trees, for instance, the Hoeffding Adaptive Trees (HAT)  \cite{Bifet:2009}. These models build trees incrementally based on the Hoeffding principle, that splits a node only when there is statistical significance between the current best attribute and the others. Later, ensemble techniques have been successfully applied to data stream classification, enhancing the predictive performance of single classifiers. ADWIN bagging used adaptive windows to control the adaptation of ensemble members to the evolution of the stream \cite{Bifet:2009}. More recently, researchers have focused on building ensemble models that can deal effectively with concept drifts. The Adaptive Random Forest (ARF) algorithm proposes better resampling methods for updating  classifiers over drifting data streams  \cite{Gomes:2017}. In \cite{Cano:2019}, the authors proposed the Kappa Updated Ensemble (KUE) that uses weighted voting from a pool of classifiers with possible abstentions.

Despite the incremental learning nature of neural networks, there is little research involving DL models in the data streaming literature. Neural networks can adapt to changes in data by updating their weights with incoming instances. However, the high training time of deep networks presents challenges to adapt them to a streaming scenario in real-time. There have been proposals using simple networks such as the Multi-Layer Perceptron \cite{Ghazikhani:2014,Zhang:2018}. A deep learning framework for data streaming that uses a dual-pipeline architecture was developed in \cite{Lara:2020}. A more detailed description of the framework, which was the first using complex DL networks for data streaming, is provided in the next section.

\section{Materials and methods}

\subsection{ADLStream framework}
\begin{figure}[H]
    \centering
    \includegraphics[scale=0.6]{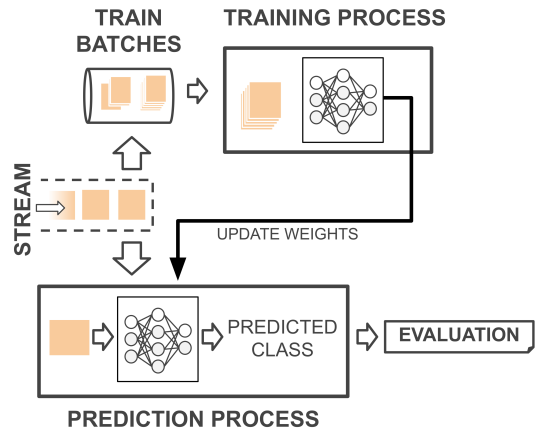}
    \caption{Asynchronous dual-pipeline deep learning framework}
    \label{fig:adlstream}
\end{figure}

In this study, we use the asynchronous dual-pipeline deep learning framework (ADLStream) for data streaming presented in \cite{Lara:2020}. As can be seen in Figure \ref{fig:adlstream}, the proposed system has two separated layers for training and predicting. This improves the processing rate of incoming data since instances are classified as soon as they arrive using a recently trained model. In the other layer, the weights of the network are constantly being updated in order to adjust to the evolution of the stream. This framework allows to use complex DL model, such as recurrent or convolutional, that would not be possible to use in a data streaming scenario if they are trained sequentially. The source code of ADLStream framework can be found at \cite{github-code}.

\subsection{Datasets}
For the experimental study, 29 one-dimensional time series datasets from the UCR repository have been simulated as streams \cite{TSA:2018}. The selected datasets have different characteristics and are categorized into six different domains. Table \ref{table:datasets} presents a detailed description of the number of instances, length of the time series instances, and the number of classes of each dataset.

\begin{table}[ht]
\centering
\caption{Datasets used for the study.}
\label{table:datasets}
\resizebox{0.9\textwidth}{!}{%
\begin{tabular}{@{}cccccc@{}}
\hline
\textbf{\#} & \textbf{Dataset} & \textbf{Instances} & \textbf{Length} & \textbf{Classes} & \textbf{Type} \\ 
\hline
1 & TwoPatterns & 5000 & 128 & 4 & SIMULATED\\
2 & CinCECGtorso & 1420 & 1639 & 4 & ECG \\
3 & TwoLeadECG & 1162 & 82 & 2 & ECG \\
4 & Wafer & 7164 & 152 & 2 & SENSOR\\
5 & Pendigits & 10992 & 16 & 10 & MOTION \\
6 & FacesUCR & 2250 & 131 & 14 & IMAGE \\
7 & Mallat & 2400 & 1024 & 8 & SIMULATED \\
8 & FaceAll & 2250 & 131 & 14 & IMAGE \\
9 & Symbols & 1020 & 398 & 6 & IMAGE \\
10 & ItalyPowerDemand & 1096 & 24 & 2 & SENSOR\\
11 & ECG5000 & 5000 & 140 & 5 & ECG \\
12 & MoteStrain & 1272 & 84 & 2 & SENSOR \\
13 & NonInvasiveFetalECGThorax1 & 3765 & 750 & 42 & ECG \\
14 & NonInvasiveFetalECGThorax2 & 3765 & 750 & 42 & ECG \\
15 & SwedishLeaf & 1125 & 128 & 15 & IMAGE \\
16 & FordA & 4921 & 500 & 2 & SENSOR \\
17 & Yoga & 3300 & 426 & 2 & IMAGE \\
18 & UWaveGestureLibraryX & 4478 & 315 & 8 & MOTION \\
19 & FordB & 4446 & 500 & 2 & SENSOR \\
20 & ElectricDevices & 16637 & 96 & 7 & DEVICE \\
21 & UWaveGestureLibraryY & 4478 & 315 & 8 & MOTION \\
22 & UWaveGestureLibraryZ & 4478 & 315 & 8 & MOTION \\
23 & HandOutlines & 1370 & 2709 & 2 & IMAGE \\
24 & InsectWingbeatSound & 2200 & 256 & 11 & SENSOR \\
25 & ShapesAll & 1200 & 512 & 60 & IMAGE \\
26 & MedicalImages & 1141 & 99 & 10 & IMAGE \\
27 & PhalangesOutlinesCorrect & 2658 & 80 & 2 & IMAGE \\
28 & ChlorineConcentration & 4307 & 166 & 3 & SIMULATED\\
29 & Phoneme & 2110 & 1024 & 39 & SENSOR \\ 
\hline
\end{tabular}
}
\end{table}

\subsection{Experimental study}
In this section, we present the design of the different types of DL models selected for the experimental study. Furthermore, we also describe the details of the evaluation method used for the data streaming classification task.

\subsubsection{Deep learning models}~\\

\begin{table}[!b]
\begin{minipage}[b]{.36\textwidth}
\centering
\caption{Multi-Layer Perceptron architecture}
\label{tab:mlp}
\begin{tabular}{ccc}
\hline
\multicolumn{3}{c}{\textbf{MLP}} \\ \hline
\textbf{Layer} & \textbf{Type} & \textbf{Neurons} \\ \hline
0 & Input & $f$  \\
1 & Dense & 32  \\
2 & Dense & 64  \\
3 & Dense & 128  \\
4 & Softmax & $c$ \\ \hline
Params & \multicolumn{2}{c}{$f \times 32 + 10240 + c \times 128$}   \\ \hline
\end{tabular}
\end{minipage}
\hspace{0.07\textwidth}
\begin{minipage}[b]{.55\textwidth}
\centering
\caption{Convolutional Neural Network architecture. $k$ indicates the kernel size}
\label{tab:cnn}
\begin{tabular}{ccc}
\hline
\multicolumn{3}{c}{\textbf{CNN}} \\ \hline
\textbf{Layer} & \textbf{Type} & \textbf{Neurons} \\ \hline
0 & Input & $f$ \\
1 & Conv. $(k=7)$ & \hspace{0.1em} $f \times 64$ maps \\
2 & Max-Pool $(k=2)$ & \hspace{0.1em} $f/2 \times 64$ maps \\
3 & Conv. $(k=5)$ & \hspace{0.1em} $f/2 \times 128$ maps \\
4 & Max-Pool $(k=2)$ & \hspace{0.1em} $f/4 \times 128$ maps \\
5 & Dense & 64 \\
6 & Dense & 32 \\
7 & Softmax & $c$ \\ \hline
Params & \multicolumn{2}{c}{$f \times 2048 + 43648 + c \times 32$}   \\ \hline
\end{tabular}%
\end{minipage}
\end{table}

\begin{table}[!b]
\begin{minipage}[b]{.4\textwidth}
\centering
\caption{Long Short-Term Memory Network architecture}
\label{tab:lstm}
\begin{tabular}{ccc}
\hline
\multicolumn{3}{c}{\textbf{LSTM}} \\ \hline
\textbf{Layer} & \textbf{Type} & \textbf{Neurons} \\ \hline
0 & Input & $f$  \\
1 & LSTM & $f \times 64$ units \\
2 & LSTM & $f \times 128$ units\\
3 & Dense & 64  \\
4 & Dense & 32  \\
5 & Softmax & $c$ \\ \hline
Params & \multicolumn{2}{c}{$f \times 8192 + 117760 + c \times 32$}   \\ \hline
\end{tabular}
\end{minipage}
\hspace{0.07\textwidth}
\begin{minipage}[b]{.45\textwidth}
\centering
\caption{Temporal Convolutional Network architecture. $k$ indicates the kernel size}
\label{tab:tcn}
\begin{tabular}{ccc}
\hline
\multicolumn{3}{c}{\textbf{TCN}} \\ \hline
\textbf{Layer} & \textbf{Type} & \textbf{Neurons} \\ \hline
0 & Input & $f$  \\
1 & TCN $(k=5)$ & $f \times 64$ maps \\
2 & Dense & 64  \\
3 & Dense & 32  \\
4 & Softmax & $c$ \\ \hline
Params & \multicolumn{2}{c}{$f \times 4096 + 372 096 + c \times 32$}   \\ \hline
\end{tabular}
\end{minipage}
\end{table}

\noindent Our aim in this study is to evaluate the performance of different DL architectures within the ADLStream framework. Four different families of architectures are considered in the experiments: the Multi-layer Perceptron (MLP) which will serve as the baseline, recurrent networks using Long Short-Term Memory cells (LSTM), Convolutional Neural Networks (CNN), and Temporal Convolutional Networks (TCN). While the MLPs is unable to model the time relationships within the input data, the last three architectures are particularly indicated for dealing with data that has a temporal or spatial grid-like structure, such as the selected datasets. LSTM networks are one of the most popular types of recurrent neural networks. They connect each time step with the previous ones in order to model the long temporal dependencies of the data without forgetting the short-term patterns using special gates \cite{Gers:2000}. On the other hand, CNNs are networks based on the convolution operation, which creates features maps using sliding filters. They are also suitable for one-dimensional time series data since they are able to automatically capture repeated patterns at different scales \cite{Krizhevsky:2012}. Moreover, they have far less trainable parameters than recurrent networks due to their weight sharing scheme \cite{Borovykh:2019}. More recently, TCNs have emerged as a specialised architecture that can capture long-term dependencies more effectively by using dilated causal convolutions. With this operation, the receptive field of neurons is increased without the need for pooling operations, hence there is no loss of resolution \cite{Yu:2016}. 
Tables \ref{tab:mlp}-\ref{tab:tcn} provide a detailed description of the layers composing the four DL models considered. In these tables, the values of $f$ and $c$ are the number of features  of the instances and the number of classes respectively. The baseline MLP model (Table \ref{tab:mlp}) is composed of three dense layers with an increasing number of neurons. As can be seen, the other three models have a similar architecture since the convolutional or recurrent layers have the same number of maps or units and are followed by fully-connected layers with the same number of neurons. In the CNN (Table \ref{tab:cnn}), two convolutional blocks with decreasing kernel size and max-pooling of stride 2 are applied before the dense layers. In the LSTM and TCN layers, the complete sequences are returned and connected to the next layers in order to use the information of all patterns extracted at different scales. In the TCN (Table \ref{tab:tcn}), only one stack of residual blocks is used, and the dilated convolution is used with kernel $(k=5)$ and dilations ($d=\{1,2,4,8,16,32,64\}$). %This results in a receptive field for each neuron of $(rb \times k \times max(d)= 1 \times 5 \times 64 = 320)$. 
Another important element to consider is the use of a dropout with rate 0.2 on all dense layers in all models, with the aim of reducing over-fitting issues. The number of trainable parameters illustrates the computational cost of each model. The TCN has the highest number, which can be 37 times greater than the MLP model.
% Receptive field = nb_stacks_of_residuals_blocks * kernel_size * last_dilation.

\subsubsection{Evaluation}~\\

\noindent For evaluating the results we use the prequential method with decaying factors, that incrementally updates the accuracy by testing the model with unseen examples \cite{Gama:2013}. The decaying factors are used as a forgetting mechanism to give more importance to recent instances for estimating the error, given the evolving nature of the stream. In our study, we use a decaying factor of $\alpha=0.99$. The process of calculating the prequential accuracy can be formulated as follows, where $L$ is the loss function and $o$ and $y$ are the real and expected output respectively. 
\begin{equation}
\label{eq:prequential-error}
P_{\alpha}(i) = \frac{\sum^{i}_{k=1} \alpha^{i-k} L(y_{k},o_{k})}{\sum^{i}_{k=1}\alpha^{i-k}} = \\ =  L(y_{i},o_{i}) + \frac{1}{\alpha} P_{\alpha}(i-1)
\end{equation}

The metric selected is the Kappa statistic, that is more suitable than standard accuracy in data streaming due to the frequent changes in the class distribution of incoming instances \cite{Bifet:2015}.  The Kappa value can be computed as shown in the following equation, where $p_0$ is the prequential accuracy and $p_c$ is the hypothetical probability of chance agreement.

\begin{equation}
\label{eq:kappa}
k = \frac{p_0 - p_c}{1-p_c}
\end{equation}

\section{Experimental results}
This section presents the Kappa accuracy results and the statistical analysis. The experiments have been carried out with an Intel Core i7-770K and two NVIDIA GeForce GTX 1080 8GB GPU. The Apache Kafka server is used to reproduce the streaming scenario since it is the most efficient tool available \cite{JuanAntonio:2017}.

\subsection{Prequential Kappa}

Table \ref{tab:kappa} presents the prequential kappa accuracy results obtained with the different models for each dataset. As can be seen, the CNN achieves the best performance for almost all the datasets considered, obtaining the highest average kappa accuracy value. The second model on average is the TCN, but closely followed by the LSTM that shows a similar performance. In general, the results prove that the ADLStream framework is able to achieve reliable results regardless of the deep learning architecture chosen. 

\subsection{Computation time analysis}

In a data streaming environment, it is fundamental to analyse the efficiency of the architectures considered. The average processing rate of each model (average time to process each incoming instance) is provided at the end of Table \ref{tab:kappa}. Logically, the MLP is the fastest model given its simple architecture. The second fastest model is the CNN, which has a significantly smaller number of parameters than the other two DL architectures. Thanks to the properties of parameter sharing, the CNN is able to process instances three times faster than the LSTM. The TCN is a more complex model with more convolutions which results in a processing rate of almost 8 times slower than the CNN.

\begin{table}[ht]
\centering
\caption{Prequential kappa accuracy results}
\label{tab:kappa}
\begin{tabular}{cccccc}
\hline
\textbf{\#} & \textbf{Dataset} & \textbf{MLP} & \textbf{LSTM} & \textbf{CNN} & \textbf{TCN} \\ \hline
1 & TwoPatterns & 0.818 & 0.999 & \textbf{1.000} & 0.999 \\
2 & CinCECGtorso & 0.348 & \textbf{0.990} & 0.994 & 0.933 \\
3 & TwoLeadECG & 0.947 & 0.941 & \textbf{0.991} & 0.987 \\
4 & Wafer & 0.581 & 0.995 & \textbf{0.996} & 0.710 \\
5 & pendigits & 0.728 & 0.987 & \textbf{0.992} & 0.953 \\
6 & FacesUCR & 0.834 & 0.952 & \textbf{0.974} & 0.952 \\
7 & Mallat & 0.963 & 0.920 & \textbf{0.986} & 0.978 \\
8 & FaceAll & 0.841 & 0.953 & \textbf{0.962} & 0.948 \\
9 & Symbols & 0.877 & 0.900 & \textbf{0.949} & 0.914 \\
10 & ItalyPowerDemand & \textbf{0.942} & 0.921 & 0.935 & 0.934 \\
11 & ECG5000 & 0.881 & \textbf{0.891} & 0.888 & 0.890 \\
12 & MoteStrain & 0.778 & 0.843 & \textbf{0.878} & 0.851 \\
13 & NonInvasiveFetalECGThorax1 & 0.851 & 0.862 & \textbf{0.881} & 0.873 \\
14 & NonInvasiveFetalECGThorax2 & 0.894 & 0.893 & \textbf{0.901} & 0.900 \\
15 & SwedishLeaf & 0.679 & 0.775 & \textbf{0.874} & 0.844 \\
16 & FordA & -0.021 & \textbf{0.691} & 0.644 & 0.632 \\
17 & Yoga & 0.213 & 0.659 & \textbf{0.737} & 0.689 \\
18 & UWaveGestureLibraryX & 0.560 & 0.748 & \textbf{0.761} & 0.732 \\
19 & FordB & 0.009 & \textbf{0.654} & 0.626 & 0.219 \\
20 & ElectricDevices & 0.353 & \textbf{0.803} & 0.801 & 0.755 \\
21 & UWaveGestureLibraryY & 0.506 & 0.641 & \textbf{0.648} & 0.608 \\
22 & UWaveGestureLibraryZ & 0.496 & 0.646 & \textbf{0.658} & 0.613 \\
23 & HandOutlines & 0.643 & 0.674 & \textbf{0.721} & 0.714 \\
24 & InsectWingbeatSound & 0.605 & 0.602 & \textbf{0.613} & 0.598 \\
25 & ShapesAll & 0.542 & 0.598 & \textbf{0.606} & 0.596 \\
26 & MedicalImages & 0.304 & 0.566 & \textbf{0.580} & 0.503 \\
27 & PhalangesOutlinesCorrect & 0.153 & 0.156 & \textbf{0.474} & 0.451 \\
28 & ChlorineConcentration & 0.242 & 0.157 & \textbf{0.900} & 0.891 \\
29 & Phoneme & 0.032 & 0.117 & \textbf{0.182} & 0.138 \\ \hline
 & \textbf{Average kappa} & \textbf{0.572} & \textbf{0.743} & \textbf{0.798} & \textbf{0.752} \\ \hline
 & \textbf{Average time per instance (ms)} & \textbf{4.993} & \textbf{22.09} & \textbf{7.347} & \textbf{47.34} \\ \hline
\end{tabular}
\end{table}

\subsection{Statistical analysis}
The ranking of the accuracy of the models obtained with the Friedman test is presented in Table \ref{tab:friedman}. The CNN model leads the ranking, with a high difference in score with respect to the rest of the models. The TCN and LSTM obtain a similar score, while the MLP offers the worst performance. The null hypothesis is rejected since the p-value obtained (\textless 0.001) is below the significance level $(\alpha=0.05)$. 

In Bergmann-Hommel's post-hoc analysis, we perform pair-wise comparisons between all models. Table \ref{tab:post-hoc} reports the p-values and conclusions obtained. As can be seen, for the CNN all null hypothesis can be rejected since the p-values are always below the significance level. Therefore, it can be concluded that there is a statistical significance in the differences between the performance of the CNN and the other architectures considered. Nevertheless, there are no significant differences between the accuracy of LSTM and the TCN.

\begin{table}[ht]
\begin{minipage}[b]{.42\textwidth}
\caption{Friedman Test Ranking}
\label{tab:friedman}
\centering
\begin{tabular}{cc}
\hline
\multicolumn{2}{c}{\textbf{Friedman Test Ranking}} \\ \hline
\textbf{CNN} & 1.200 \\
TCN & 2.533 \\
LSTM & 2.566 \\
MLP & 3.700 \\ \hline
\end{tabular}
\end{minipage}
\begin{minipage}[b]{.5\textwidth}
\caption{Bergmann-Hommel's analysis}
\label{tab:post-hoc}
\centering
\begin{tabular}{lccc}
\hline
\multicolumn{4}{c}{\textbf{PostHoc Analysis}} \\ \hline
\textbf{Comparison} & p & z &   Conclusion \\
MLP - CNN & \textless  0.001 & 7.5 & !=  \\
LSTM - CNN &  \textless  0.001 & 4.1 & !=  \\
TCN  - CNN &  \textless  0.001 & 4 &   !=\\
MLP - TCN &  \hspace{0.2cm}  0.001 & 3.49 & !=   \\
MLP - LSTM &  \hspace{0.2cm}  0.001 & 3.39 & !=  \\
LSTM - TCN &  \hspace{0.2cm}  0.920 & 0.09 &  == \\
\hline
\end{tabular}
\end{minipage}
\end{table}

\clearpage
\section{Conclusions}
In this paper, the performance of several deep learning architectures for data streaming classification is compared using the ADLStream framework. An extensive study over a large number of time-series dataset was conducted using multi-layer perceptron, recurrent, and convolutional neural networks.

The research carried out for this study provided evidence that convolutional neural networks are currently the most suitable model for time series classification in streaming. Convolutional neural networks obtained the best results in terms of accuracy, with a very high processing rate. These characteristics present convolutional networks as the best alternative for processing data arriving at high speed. The other deep models, such as Long Short-Term Memory or Temporal Convolutional networks were not able to achieve such performance and their processing rate was slower.

Future work should study the behaviour of different deep learning models over concept drifts and their capacity to adapt to changes in the data distribution. Furthermore, a parameter optimization process could provide more specific architectures for the models and improve the performance. Future studies should also consider other less known models such as Echo State Networks, Stochastic Temporal Convolutional Networks or Gated Recurrent Units Networks.

\section*{Acknowledgements}
{This research has been funded by the Spanish Ministry of Economy and Competitiveness under the project TIN2017-88209-C2-2-R and by the Andalusian Regional Government under the projects: BIDASGRI:  Big~Data technologies for Smart Grids (US-1263341), Adaptive hybrid models to predict solar and wind renewable energy production (P18-RT-2778). We are grateful to NVIDIA for their GPU Grant Program that has provided us high quality GPU devices for carrying out the study.}


\bibliographystyle{splncs04}
\bibliography{paper.bib}

\end{document}